\documentclass[runningheads,a4paper]{llncs}

\usepackage{amssymb}
\usepackage{graphicx}

\usepackage{amsmath}
\usepackage{times}
\usepackage{soul}
\usepackage{url}
\usepackage[colorlinks=true,bookmarks=false]{hyperref} 
\usepackage[utf8]{inputenc}
\usepackage[small]{caption}
\usepackage{graphicx,subfigure}
\usepackage{graphicx}
\usepackage{algorithm}
\usepackage{algorithmic}

\usepackage{url}

\begin{document}

\mainmatter  

\title{M-ADDA: Unsupervised Domain Adaptation\\ with Deep Metric Learning}

\titlerunning{M-ADDA: Unsupervised Domain Adaptation with Deep Metric Learning}

%
%
\author{Issam H. Laradji%
\and Reza Babanezhad}
%

\institute{Department of Computer Science, University of British Columbia\\
Vancouver, British Columbia, Canada\\
\{issamou,rezababa\}@cs.ubc.ca}

\maketitle
\begin{abstract}
Unsupervised domain adaptation techniques have been successful for a wide range of problems where supervised labels are limited. The task is to classify an unlabeled `target' dataset by leveraging a labeled `source' dataset that comes from a slightly similar distribution. We propose metric-based adversarial discriminative domain adaptation (M-ADDA) which performs two main steps. First, it uses a metric learning approach to train the source model on the source dataset by optimizing the triplet loss function. This results in clusters where embeddings of the same label are close to each other and those with different labels are far from one another. Next, it uses the adversarial approach (as that used in ADDA \cite{2017arXiv170205464T}) to make the extracted features from the source and target datasets indistinguishable. Simultaneously, we optimize a novel loss function that encourages the target dataset's embeddings to form clusters. While ADDA and M-ADDA use similar architectures, we show that M-ADDA performs significantly better on the digits adaptation datasets of MNIST and USPS. This suggests that using metric-learning for domain adaptation can lead to large improvements in classification accuracy for the domain adaptation task. The code is available at \url{https://github.com/IssamLaradji/M-ADDA}.
\end{abstract}

\section{Introduction}
\label{sec:intro}

Convolutional neural networks (CNN) \cite{lecun1998gradient} allow us to extract powerful features that can be used for tasks such as image classification and segmentation. However, these features are usually domain specific in that they are not discriminative enough for datasets coming from other domains, resulting in poor classification performance. Consequently, unsupervised domain adaptation techniques have emerged \cite{ganin2016domain,tzeng2015simultaneous,liu2016coupled,wang2018deep} to address the domain shift phenomenon between a source dataset and a target dataset. Common techniques use adversarial learning in order to make extracted features from the source and target datasets indistinguishable. The extracted features from the target dataset are then passed through a trained classifier (pre-trained on the source dataset) to predict the labels of the target test-set \cite{2017arXiv170205464T}. 

\begin{figure}
\centering     
\includegraphics[width=0.5\textwidth]{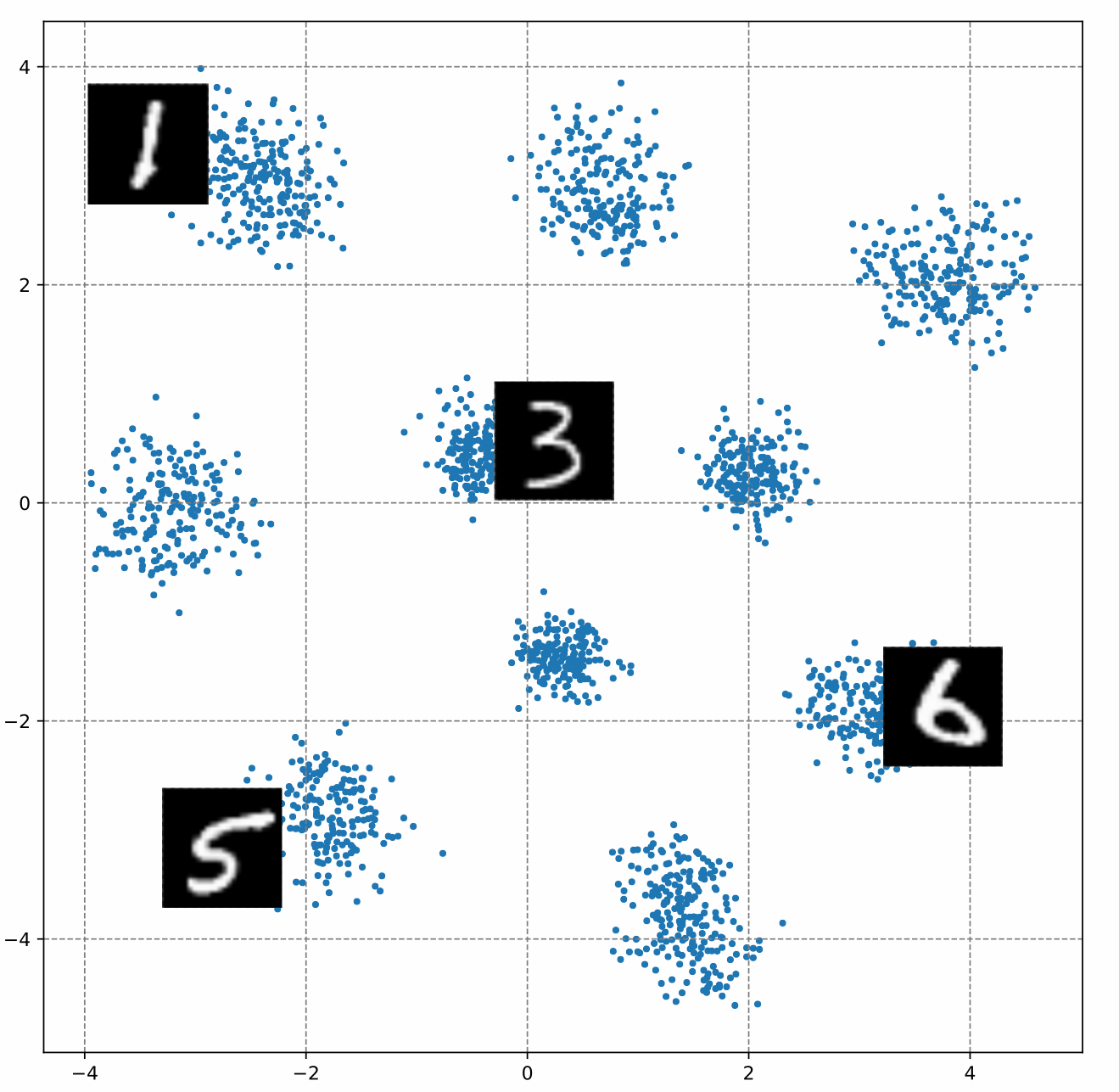}
\caption{{\bf Metric Learning.} The result of minimizing the triplet loss on the MNIST dataset. Each cluster corresponds to examples belonging to a single digit label.}
\label{fig:clusters}
\end{figure}

Recently, metric-based methods were introduced to address the problem of unsupervised domain adaptation \cite{hsu2017learning,pinheiro2017unsupervised}. Namely, classifying an example is performed by computing its similarity to prototype representations of each category \cite{pinheiro2017unsupervised}. Further, a category-agnostic clustering network was proposed by \cite{hsu2017learning} to cluster new datasets through transfer learning.
\begin{figure*}[t!]
\centering     
\subfigure{\label{fig:a}\includegraphics[width=0.45\textwidth]{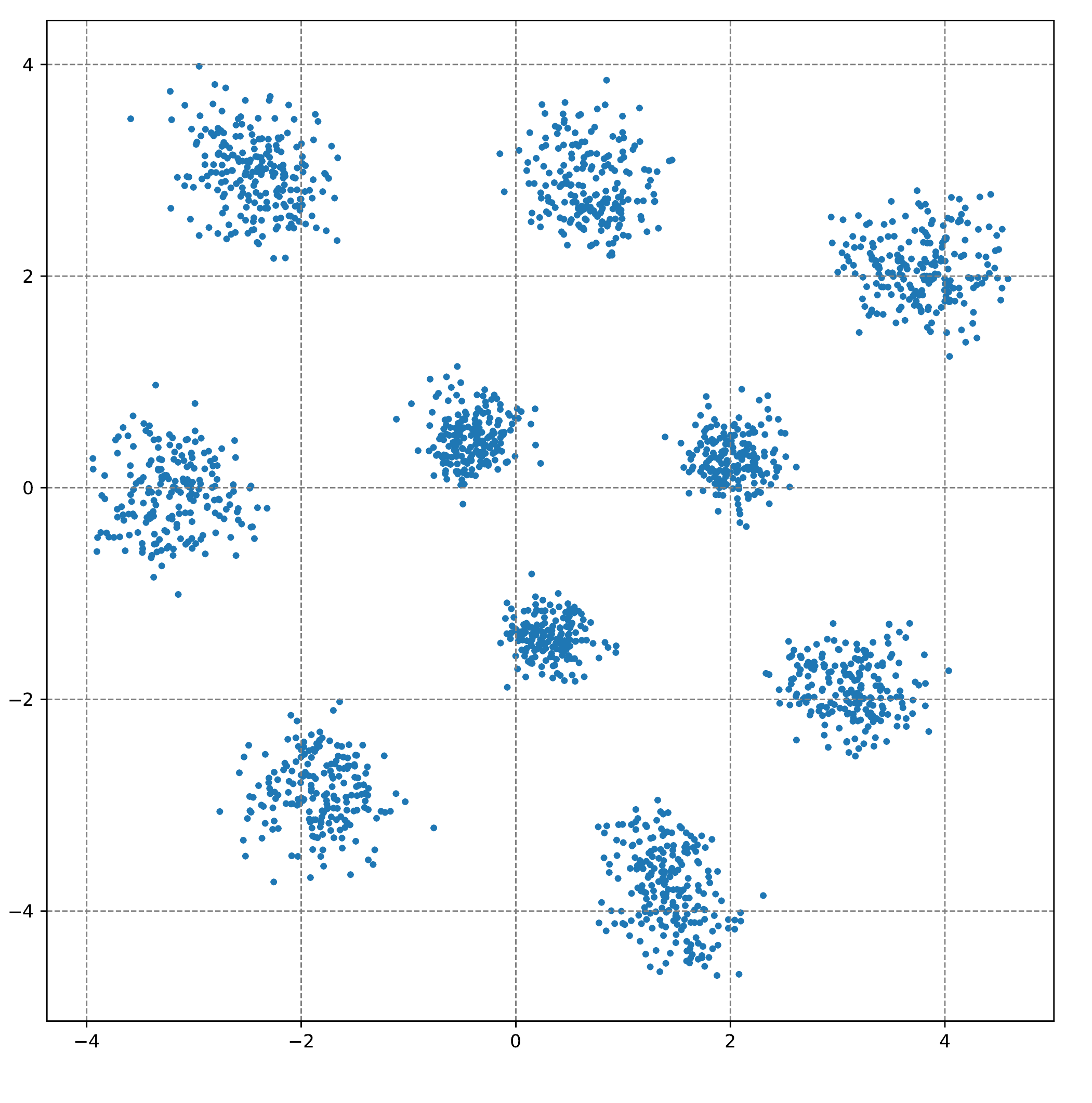}}
\subfigure{\label{fig:b}\includegraphics[width=0.45\textwidth]{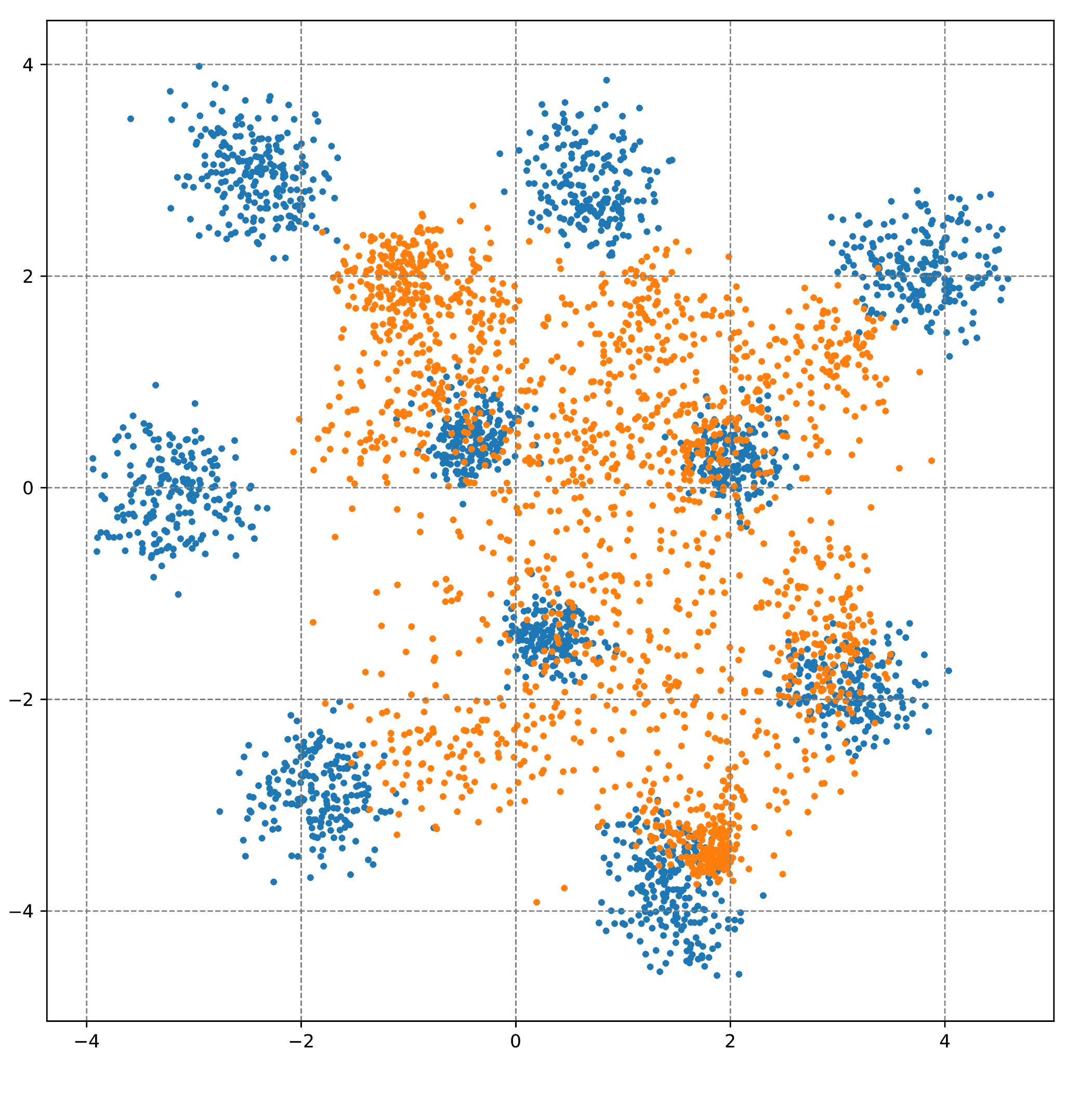}}
\caption{{\bf Domain Adaptation.} The {\bf blue} dots represent the MNIST embeddings after optimizing Eq. \eqref{eq:triplet}. The {\bf orange} dots represent the USPS embeddings. The center image shows the USPS embeddings before minimizing the domain shift adverbially by Eq. \eqref{eq:da}. The right-most image shows the USPS embeddings after optimizing Eq. \eqref{eq:dacluster}.}
\end{figure*}
In this paper, we introduce M-ADDA, a metric-based adversarial discriminative domain adaptation framework. First, M-ADDA trains our source model using metric learning by optimizing the triplet loss \cite{hoffer2015deep} on the source dataset. As a result, if $K$ is the number of classes then the dataset is clustered into $K$ clusters where each cluster is composed of examples having the same label (see Fig. \ref{fig:clusters}). The goal is to obtain an embedding of the target dataset where the k-nearest neighbors (kNN) of each example belong to the same class and where examples from different classes are separated by a large margin. A major strength in this approach is its non-parametric nature \cite{weinberger2009distance} as it does not implicitly make parametric (possibly limiting) assumptions about the input distributions.

\begin{figure*}[t!]
\centering     
\subfigure{\label{fig:a}\includegraphics[width=0.45\textwidth]{figures/p1_out.pdf}}
\subfigure{\label{fig:c}\includegraphics[width=0.45\textwidth]{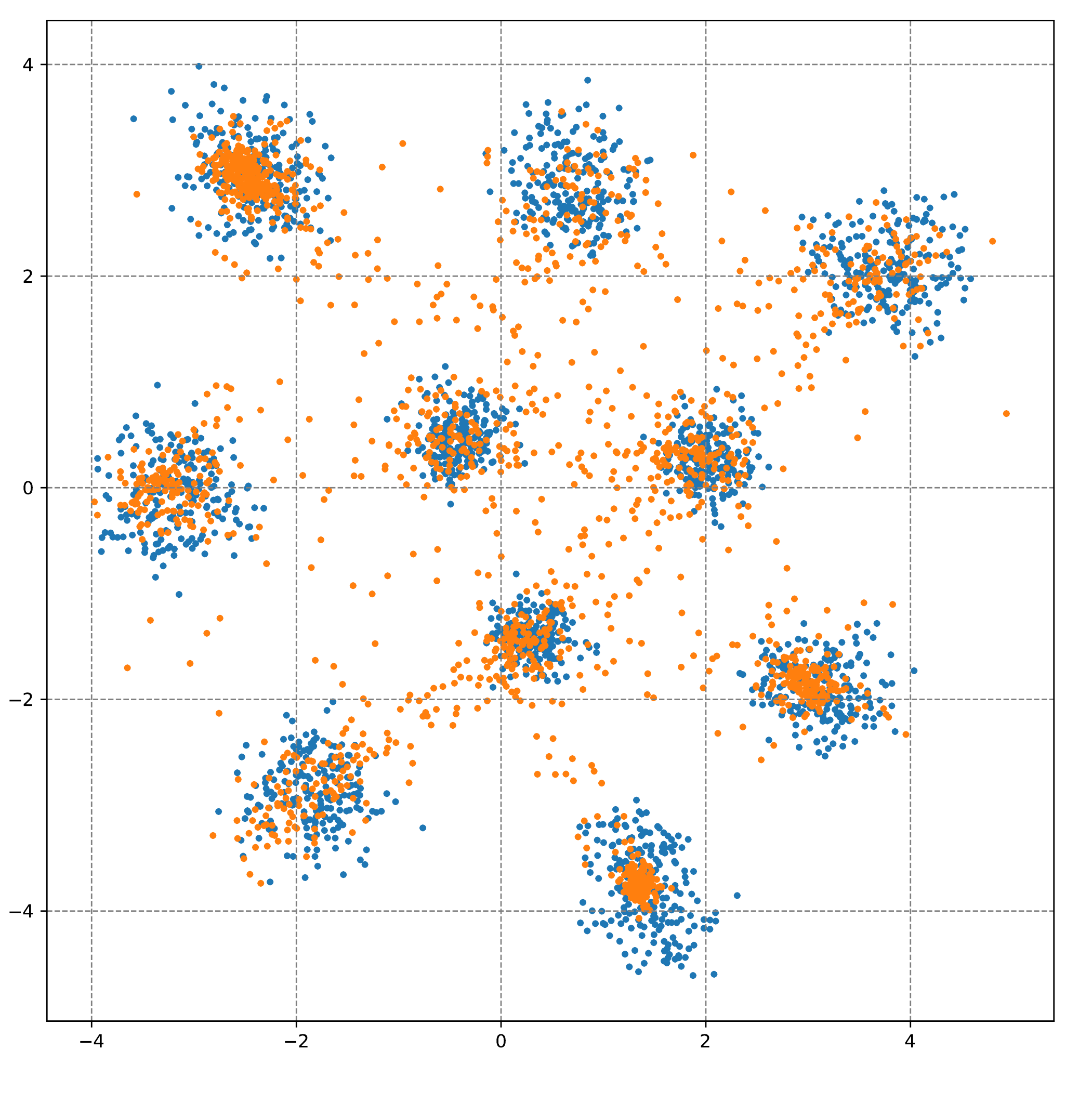}}
\caption{{\bf Domain Adaptation.} The {\bf blue} dots represent the MNIST embeddings after optimizing Eq. \eqref{eq:triplet}. The {\bf orange} dots represent the USPS embeddings. The center image shows the USPS embeddings before minimizing the domain shift adverbially by Eq. \eqref{eq:da}. The right-most image shows the USPS embeddings after optimizing Eq. \eqref{eq:dacluster}.}
\end{figure*}

Next we adapt the distributions between the source and target extracted features using the adversarial learning method used by ADDA \cite{2017arXiv170205464T}. This addresses the domain discrepancy between the datasets. Early methods for domain adaptation are based on minimizing correlation distances and minimizing the maximum mean discrepancy to ensure both datasets have a common feature space~\cite{tzeng2014deep,long2015learning,sun2016return,sun2016deep}. However, adversarial learning approaches showed state-of-the-art performance for domain adaptation. While the features' distributions become more similar during training, we also train a network that maps the extracted features to embeddings such that they are clustered into $K$ clusters. Concurrently, we encourage the clusters to have large margins between them. Therefore, the network is trained by minimizing the distance between each target example embedding and its closest cluster center corresponding to the source embedding. This approach is simple to implement and achieves competitive results on digit datasets such as MNIST \cite{lecun1998mnist}, and USPS \cite{le1989handwritten}.

To summarize our contributions, (1) we propose a novel metric-learning framework that uses the triplet loss to cluster the source dataset for the task of domain adaptation; (2) we propose a new loss function that regularizes the embeddings of the target dataset to encourage them to form clusters; and (3) we show a large improvement over ADDA \cite{2017arXiv170205464T} on a standard unsupervised domain adaptation benchmark. Note that ADDA uses a similar architecture but a different loss function than M-ADDA.

In section \ref{sec:related}, we review the related works and other similar approaches. In section \ref{sec:proposed}, we introduce our framework and the new loss terms for domain adaptation. In section  \ref{sec:exps}, we present experimental results illustrating the efficacy of our approach on the digits dataset. Finally, we conclude the paper in section \ref{sec:conclusion}.

\begin{figure*}
\centering     
\subfigure{\label{fig:a}\includegraphics[width=1.0\textwidth]{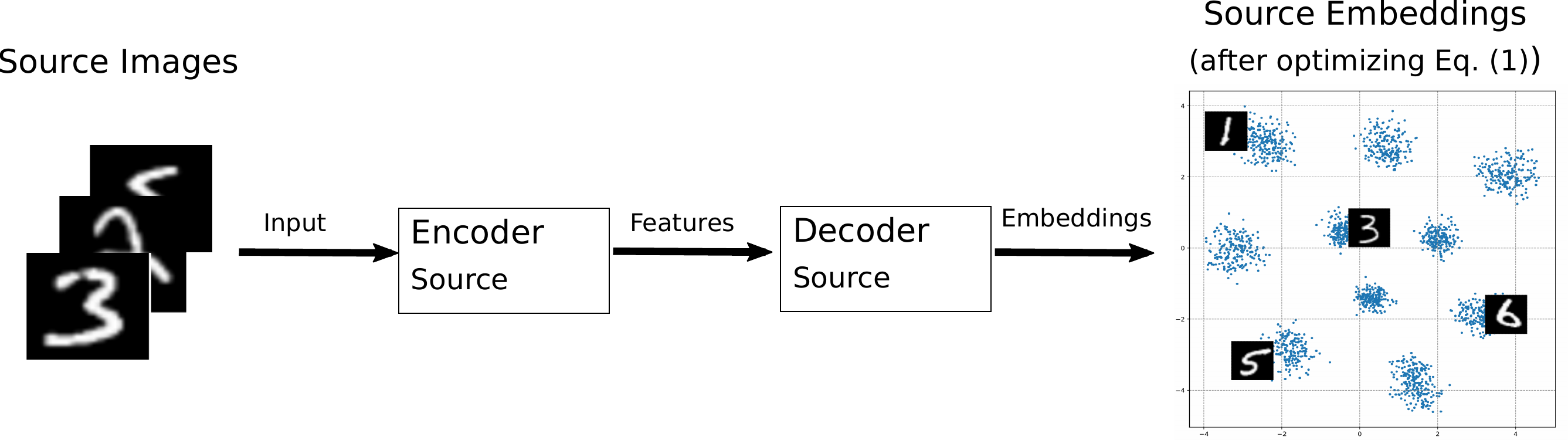}}
\caption{{\bf Training the source model.} We pre-train the source encoder and decoder by optimizing the triplet loss in Eq. \eqref{eq:triplet}. The source encoder extracts the features from the source dataset and the decoder maps the features to the embedding space where clusters are formed.}
\label{fig:pretrain}
\end{figure*}

\section{Related Work}\label{sec:related}

{\it Metric learning} has shown great success in many visual classification tasks \cite{weinberger2009distance,song2016deep,hoffer2015deep}. The goal is to learn a distance metric such that examples belonging to the same label are close as possible in some embedding space and samples from different labels are as far from one another as possible. It can be used for unsupervised learning such as clustering \cite{xing2003distance} and supervised learning such as k-nearest neighbor algorithms \cite{han2001text,weinberger2009distance}. Recently, triplet networks \cite{hoffer2015deep} and Siamese networks \cite{bertinetto2016fully} were proposed as powerful models for metric learning which have been successfully applied for few-shot learning and learning with few data. However, to the best of our knowledge we are the first to apply metric learning that is based on triplet networks for domain adaptation.

A close topic to domain adaptation is {\it transfer learning} which has received tremendous attention recently. It allows us to solve tasks where labels are scarce by learning from relevant tasks for which labels are abundant \cite{cao2013practical,shi2017transfer,deselaers2012weakly} by identifying a common structure between multiple tasks \cite{finn2017model}. A common transfer learning strategy is to use pre-trained networks such as those trained on imagenet \cite{krizhevsky2012imagenet} and fine-tune them on new tasks. While this approach can significantly improve performance for many visual tasks, it performs poorly when the pre-trained network is used on a dataset which comes from a different distribution than the one it trained on. This is because the model has learned features that are specific to one domain that might not be meaningful for other domains. 

To address this challenge, a large set of domain adaptation methods were proposed over the years \cite{2017arXiv170205464T,ganin2016domain,tzeng2015simultaneous,liu2016coupled} whose goal is to determine a common latent space between two domains often referred to as a source dataset and a target dataset. The general setting is to use a model that trains to extract features from the source dataset, and then encourage features extracted from the target dataset to be similar to the source features \cite{ghifary2016deep,bousmalis2016domain,ganin2014unsupervised,tzeng2017adversarial,saito2017asymmetric}. Auto-encoder based methods~\cite{ghifary2016deep,bousmalis2017unsupervised} train one or a variety of auto-encoders for the source and target datasets. Then, a classifier is trained based on the latent representation of the source dataset. The same classifier is then used to label the target dataset. Adversarial networks~\cite{goodfellow2014generative} based approaches use a generator model to transform the examples' feature representations from one domain to another~\cite{bousmalis2017unsupervised,shrivastava2017learning,russo2017source}. 

Another group of domain adaptation methods~\cite{tzeng2017adversarial,tzeng2014deep,li2016revisiting,sun2016deep} minimize the difference between the distributions of the features extracted from the source and target data. They achieve this by minimizing point estimates of a given metric between the source and target distributions by using maximum or mean discrepancy metrics. Current state-of-the-art techniques use the adversarial learning approach to encourage the feature representations from the two datasets to be indistinguishable (i.e. have a common distribution) \cite{2017arXiv170205464T}. Close to our method are the recent similarity based approaches proposed by \cite{hsu2017learning,pinheiro2017unsupervised}, which transfer class-agnostic prior to new datasets, and classify examples by computing their similarity to prototype representation of each category, respectively. Our approach uses a regularized metric learning method with the help of k-nearest neighbors as a non-parametric framework. This can be more powerful than ADDA which uses a model that makes parametric assumptions (introducing limitations) about the input distribution \cite{weinberger2009distance}.

Another class of domain adaptation methods are self-ensembling methods which augment the source dataset by applying various label preserving transformations on the images~\cite{laine2016temporal,tarvainen2017mean,french2018self,sajjadi2016regularization}. Using the augmented dataset they train several deep network models and use an ensemble of those networks for the domain adaptation task. Laine et. al.~\cite{laine2016temporal} have two networks in their model: the $\Pi$-model and temporal model. In the $Pi$-model, every unlabelled sample feeds to a classifier twice with different dropout, noise and image translation parameters. Their temporal model records the average of the historical network prediction per sample and forces the subsequent predictions to be close to the average. Travainen et.al~\cite{tarvainen2017mean} improve the temporal network by recording the average of the network weights rather than class prediction. This results in two networks: the student and the teacher network. The student network is trained via gradient descent and the weights of the teacher are the historical exponential moving average of the weights of the student network. The unsupervised loss is the mean square difference between the prediction of the student and the teacher under different dropout, noise and image translation parameters. French et. al.~\cite{french2018self} combine the previous two methods with adding extra modifications and engineering and gets state of the art results in many domain adaptation tasks for image datasets. However, this method uses heavy engineering with many label preserving transformations to augment the data. In contrast, we show that our method significantly improves results over ADDA by making simple changes to their framework.

\section{Proposed Approach: M-ADDA}\label{sec:proposed}

\begin{figure*}
\centering     
\subfigure{\label{fig:a}\includegraphics[width=1.0\textwidth]{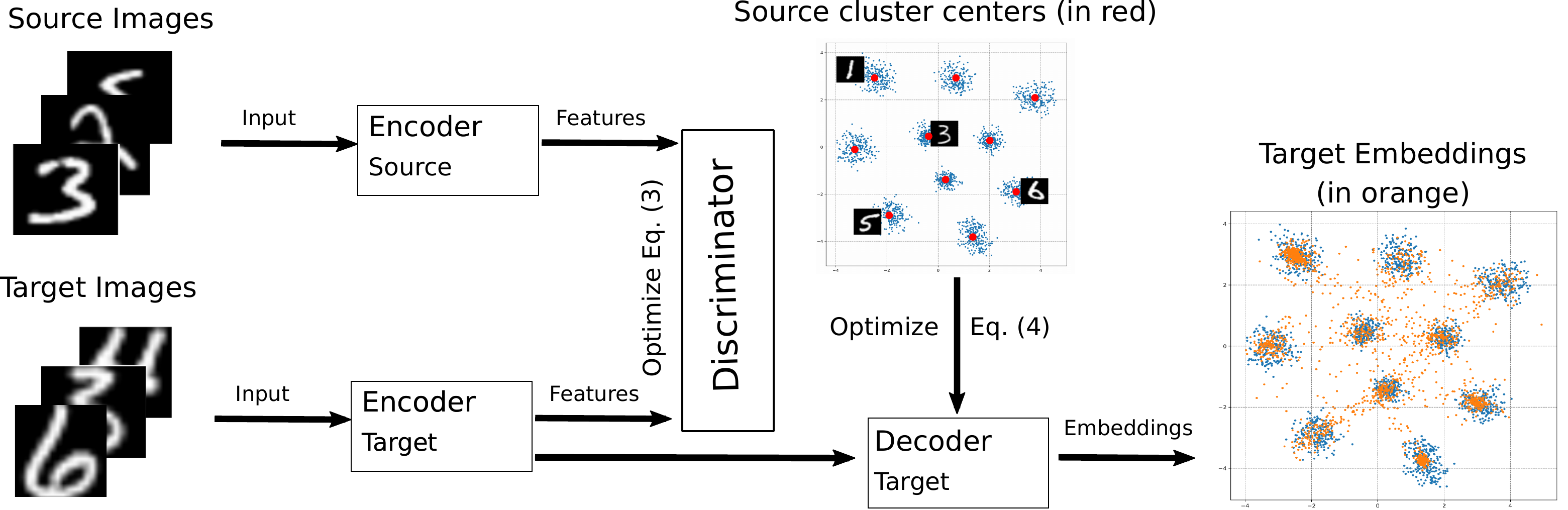}}
\caption{{\bf Training the target model.} We  adversarially adapt the encoded features' distributions between the source and target encoder using Eq. \eqref{eq:da} while using the source cluster centers to optimize Eq. \eqref{eq:cluster}. The label of each target embedding is the mode of the labels of the nearest source embedding neighbors.}
\label{fig:adversarial}
\end{figure*}

We propose M-ADDA which performs two main steps:
\begin{enumerate}
    \item train a source model on the source dataset using metric learning (as in Figure \ref{fig:pretrain}) using the Triplet loss function; then
    \item simultaneously, adapt the distributions between the extracted source and target dataset features and regularize the predicted target dataset embeddings to form clusters (see Figure \ref{fig:adversarial}).
\end{enumerate}

Our M-ADDA framework consists of a source model and a target model. The two models have the same architecture, and they both have an encoder that extracts features from the input dataset and a decoder to map the extracted features to embeddings. Consider a source dataset $(X_S, Y_S)$, and a target dataset $(X_T, Y_T)$ where the data $X_S$ and $X_T$ are drawn from two different distributions. 

\subsubsection{Training the source model.}
The source model $f_{\theta_S}(\cdot)$, parameterised by $\theta_S$, is first trained on the source dataset by optimizing the following triplet loss:
\begin{equation}
\begin{aligned}
\mathcal{L}(\theta_S)= \sum_{(a_i, p_i, n_i)}\max(&||f_{\theta_S}(a_i) - f_{\theta_S}(p_i)||^2 - \\ & ||f_{\theta_S}(a_i) - f_{\theta_S}(n_i)||^2 + m,0)
\label{eq:triplet}
\end{aligned}
\end{equation}
where $a_i$ is an anchor example (picked randomly), $p_i$ is an example with the same label as the anchor and $n_i$ is an example with a different label from the anchor. Optimizing Eq.~\eqref{eq:triplet} encourages the embedding of $a_i$ to be closer to $p_i$ than to $n_i$ by at least margin $m$. If the anchor example is close enough to the positive example $p_i$, and far from the negative example $n_i$ by a margin of at least $m$, the $max$ function returns zero; therefore, the corresponding triplet $(a_i,p_i,n_i)$ does not contribute to the loss function. If the margin is smaller than $m$, then the $max$ function returns $||f_{\theta_S}(a_i) - f_{\theta_S}(p_i)||^2 - ||f_{\theta_S}(a_i) - f_{\theta_S}(n_i)||^2 + m$. Minimizing this term results in moving $a_i$ towards $p_i$ and moving it away from $n_i$ in the embedding feature space. After optimizing the loss term long enough, the samples with the same label are pulled together and those with different labels are pushed away from each other. As a result, points of the same label form a single cluster which allows us to efficiently classify examples using k-nearest neighbors (see Figure \ref{fig:pretrain}). 

Algorithm \ref{alg:source} shows the procedure of training the source model on the source dataset for one epoch.  Given a batch $(X_B, Y_B)$, for each unique element $y_i$ in $Y_B$, we obtain an anchor $a_i$ whose label is $y_i$, a positive example $p_i$ whose label is $y_i$, and a negative example $n_i$ whose label is not $y_i$. Note that set($Y_B$) returns the unique elements of $Y_B$. In our experiments, we obtained the negative example uniformly at random. However, other methods are possible such as greedily picking the triplet with the largest loss (as computed by Eq. (\ref{eq:triplet})), and non-uniformly picking triplets based on their individual loss values. Finally, for each triplet, we compute the loss and update the parameters of the source model to minimize Eq. (\ref{eq:triplet}). 

\subsubsection{Training the target model.} Next, we define $C$ as the set of centers corresponding to the source embedding clusters (represented as red dots in Figure \ref{fig:adversarial}). Each center in $C$ corresponds to a single label in the source dataset. A center is computed by taking the mean of the source embeddings belonging to that center's label. Then, we train the target model, parametrized by $\theta_T$ by optimizing the following two loss terms:
\begin{equation}
\mathcal{L}(\theta_T, \theta_D) =  \underbrace{\mathcal{L}_A(\theta_{T_E}, \theta_{D})}_{\text{Adapt}} + \underbrace{\mathcal{L}_C(\theta_{T})}_{\text{C-Magnet}}
\label{eq:dacluster}
\end{equation}
where $\theta_{T_E}$ correspond to the parameters of the target model's encoder; and $\theta_D$ is the parameter set for a discriminator model we use to adapt the distributions of the extracted features between the source ($S$) and target ($T$) datasets. We achieve this by optimizing:

\begin{equation}
\begin{aligned}
\mathcal{L}_A(\theta_{T_E}, \theta_D)= \min_{\theta_{D}} \max_{\theta_{T_E}} - &\sum_{i\in S} \log{D_{\theta_{D}}(E_{\theta_S}(X_{S_i}))} \;- \\& \sum_{i\in T} \log{(1 - D_{\theta_{D}}(E_{\theta_{T_E}}(X_{T_i})))},
\label{eq:da}
\end{aligned}
\end{equation}
where $\theta_{S_E}$ is the source model encoder's set of parameters; and  $D(\cdot)$ is the discriminator model which is trained to maximize the probability that the features extracted by the source model's encoder come from the source dataset and that the features extracted by the target model's encoder come from the target dataset. In other words, the discriminator $D(.)$ tries to distinguish between the features extracted from the source dataset and the features from the target dataset by giving higher value (close to one) to a source dataset feature vector and a lower value (close to zero) to a target dataset feature vector. Simultaneously, the encoder of the target model is trained  to confuse the discriminator into predicting the target features as coming from the source dataset. This adversarial learning approach encourages the features extracted by $E_{\theta_{S_E}}(X_{S_i})$ and $E_{\theta_{T_E}}(X_{T_i})$ to be indistinguishable in their distributions. For the sake of brevity, note that we show the loss functions in terms of a single source example $X_{S_i}$ and target example $X_{T_i}$.

In parallel, we minimize the center magnet loss term defined as,
\begin{equation}
\mathcal{L}_C(\theta_T)= \sum_{i\in T} \min_j ||f_{\theta_T}(x_i) -  C_j||^2,
\label{eq:cluster}
\end{equation}
which pulls the embeddings of example $X_i$ to the closest cluster center defined in $C$  (see Figure \ref{fig:adversarial}). The cluster center for a class is obtained by taking the Euclidean mean of all samples belonging to that class. Since we have 10 classes in MNIST and USPS, $|C|=10$. This regularization term allows the target dataset embeddings to form clusters that are similar to the clusters formed by the source dataset embeddings. This is useful when minimizing $\mathcal{L}(\theta_T, \theta_D)$ fails to make the target embedding clustered in a similar way as the source embeddings. For example, in Fig. \ref{fig:b} we see that the target embeddings become scattered around the center when minimizing $\mathcal{L}_A(\theta_T, \theta_D)$ only. However, by simultenously minimizing $\mathcal{L}_C(\theta_T)$ we get a better formation of clusters as seen in Fig. \ref{fig:c}.


\begin{algorithm}
  \caption{Training the source model on the source dataset (single epoch).}\label{alg:source}
  \begin{algorithmic}[1]
  \INPUTS
    \STATE Source model $f_{\theta_S}(\cdot)$, and source images and labels $(X_S, Y_S)$.
  \ENDINPUTS
    \FOR{$\{X_B, Y_B\} \in (X_S, Y_S)$}
    \FOR{$y_i \in \text{set( }Y_B\text{) }$}
      \STATE $AP \gets \text{All image pairs whose label is } y_i$.
      \FOR{each $\{a_i,p_i\} \in AP$ }
        \STATE $n_i \gets \text{A random sample in } X_B \text{ whose label is not } y_i$.
        \STATE $L \gets \text{The loss in Eq (\ref{eq:triplet}) using } \{a_i, p_i, n_i\} \text{ and } f_{\theta_S}(\cdot)$.
        \STATE Update the parameters $\theta_S$ by backpropagating through $L$.
      \ENDFOR
    \ENDFOR
    \ENDFOR
  \end{algorithmic}
\end{algorithm}

\begin{algorithm}
  \caption{Training the target model on the target dataset (single epoch).}\label{alg:da}
  \begin{algorithmic}[1]
  \INPUTS
    \STATE Target model $f_{\theta_T}(\cdot)$, and source and target images and labels $(X_S, Y_S, X_T, Y_T)$.
  \ENDINPUTS
    \FOR{$\{X_{S_B}, Y_{S_B},X_{T_B}, Y_{T_B}\} \in (X_S, Y_S, X_T, Y_T)$}
        \STATE Maximize Eq. (\ref{eq:da}) w.r.t. $\theta_D$ using $\{X_{S_B}, Y_{S_B},X_{T_B}, Y_{T_B}\}$ 
        \STATE Minimize Eq. (\ref{eq:da}) w.r.t. $\theta_T$ using $\{X_{S_B}, Y_{S_B},X_{T_B}, Y_{T_B}\}$ 
    \ENDFOR
    \STATE $E_S \gets \text{The embeddings of the source dataset extracted by} f_{\theta_S}(\cdot)$
    \STATE $C \gets \text{The cluster centers of } E_S \text{ are obtained by taking the Euclidean mean for each class.}$
    \FOR{$\{X_{T_B}, Y_{T_B}\} \in (X_T, Y_T)$}
      \STATE $L \gets \text{The loss computed using Eq. \ref{eq:cluster} and cluster centers $C$}$
      \STATE Update parameters $\theta_T$ by backpropagating through $L$.
      \ENDFOR
  \end{algorithmic}
\end{algorithm}

\begin{algorithm}
  \caption{Predicting the labels of the test images.}\label{alg:c}
  \begin{algorithmic}[1]
  \INPUTS
    \STATE Target model $f_{\theta_T}(\cdot)$, Source model $f_{\theta_T}(\cdot)$, and source and target images and labels.
  \ENDINPUTS
    \STATE $E_S \gets \text{The embeddings of the source dataset extracted by} f_{\theta_S}(\cdot)$
   \FOR{$\{X_{T_B}, Y_{T_B}\} \in (X_T, Y_T)$}
       \STATE $E_{T_B} \gets \text{The embeddings of $X_{T_B}$ extracted by} f_{\theta_T}(\cdot)$
       \STATE $P_{T_B} \gets \text{The mode label of the k-nearest $E_S$ samples.}$
    \ENDFOR
  \end{algorithmic}
\end{algorithm}

Algorithm \ref{alg:da}  shows the procedure for training the target model on the target dataset. Lines 4-5 use Eq. (\ref{eq:da}) to make the target features and the source features indistinguishible. Lines 7-12 update the target model parameters by encouraging the target embeddings to move to the closest source cluster center. As shown in Algorithm \ref{alg:c}, the prediction stage consists of two steps. First we extract the embeddings of the source dataset examples using the pre-trained source model. Then, the label of an example $X_{T_i}$ is the mode label of the k-nearest source embeddings. This non-parametric approach allows us to implicitly learn powerful features that are used to compute the similarities between the examples.
\section{Experiments}\label{sec:exps}

\begin{table}[t]
\centering
\caption{{\bf Digits Adaptation}. We evaluate our method on the unsupervised domain adaptation task on the digits datasets, using the setup in \cite{2017arXiv170205464T}.}
\begin{tabular}{ |l|c|c| } 
   Method & MNIST $\rightarrow$ USPS & USPS $\rightarrow$ MNIST \\\hline\hline
 Source only (ADDA \cite{2017arXiv170205464T})& 0.752 & 0.571 \\\hline
 Source only (Ours)& 0.601 &  0.679\\\hline
   Gradient reversal \cite{ganin2016domain}&  0.771  & 0.730 \\\hline
   Domain confusion \cite{tzeng2015simultaneous}& 0.791  & 0.665  \\\hline
  CoGAN  \cite{liu2016coupled}& 0.912& 0.891   \\\hline
  ADDA   \cite{2017arXiv170205464T}& 0.894 & 0.901  \\\hline\hline
M-ADDA (Ours)  & {\bf  \;0.952 } & {\bf \;0.940 } \\\hline
\end{tabular}
\label{table:results}
\end{table}

\begin{figure}[t]
\centering     
\subfigure{\label{fig:a}\includegraphics[width=70mm]{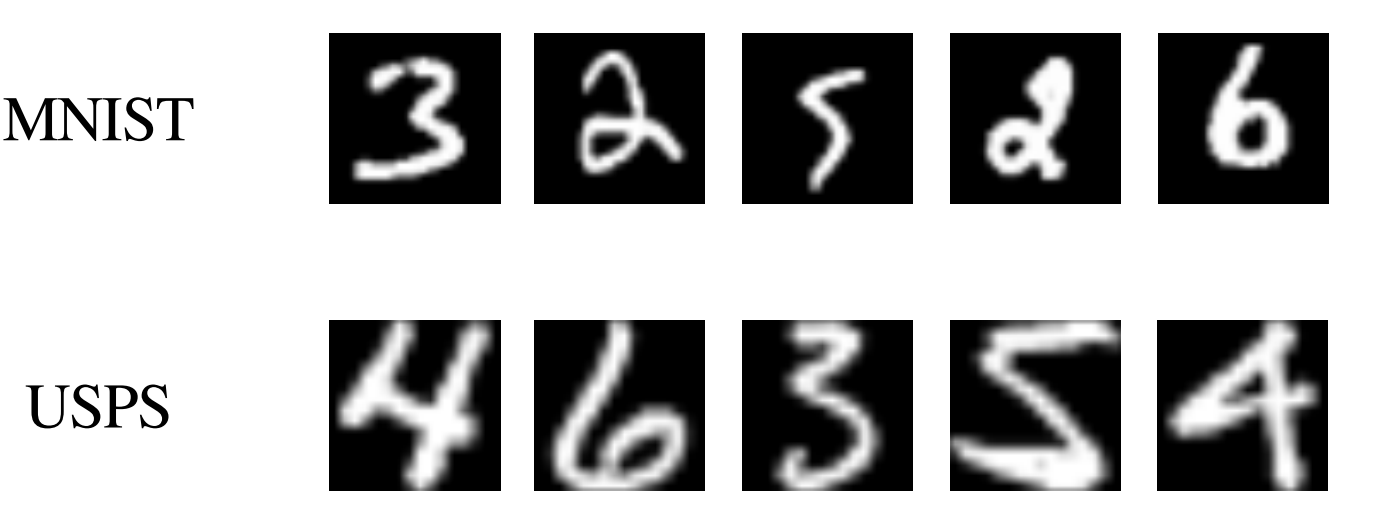}}
\caption{{\bf Dataset.} Example images taken from the 2 digit domains we used in our benchmark.}
\label{fig:digits}
\end{figure}

\begin{table}[t]
\centering
\caption{{\bf Digits Adaptation}. We evaluate our method using the setup in \cite{bousmalis2016domain,bousmalis2017unsupervised}.}
\begin{tabular}{ |l|c|c| } 
   Method & MNIST $\rightarrow$ USPS & USPS $\rightarrow$ MNIST \\\hline\hline

 Source only (Ours)& 0.60 &  0.68\\\hline
  DSN \cite{bousmalis2016domain}& 0.91 &  -\\\hline
   PixelDA \cite{bousmalis2017unsupervised} & 0.96 & -\\\hline
  SimNet   \cite{pinheiro2017unsupervised}& 0.96 & 0.96  \\\hline\hline
M-ADDA (Ours)  & {\bf  0.98} & {\bf 0.97 } \\\hline
\end{tabular}
\label{table:resultsBig}
\end{table}

\begin{figure}[t]
    \centering
        \includegraphics[width=0.5\textwidth]{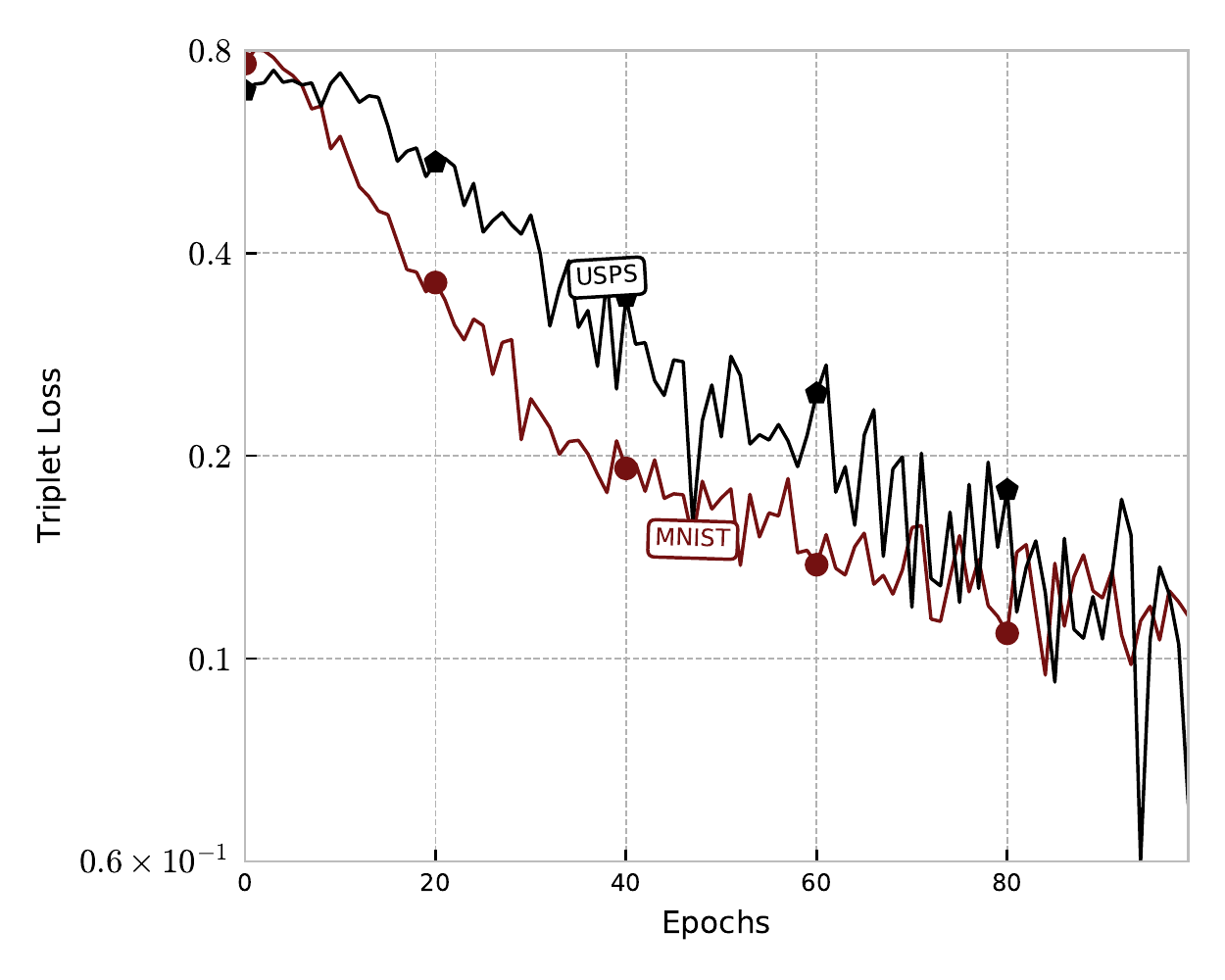}~
        \includegraphics[width=0.5\textwidth]{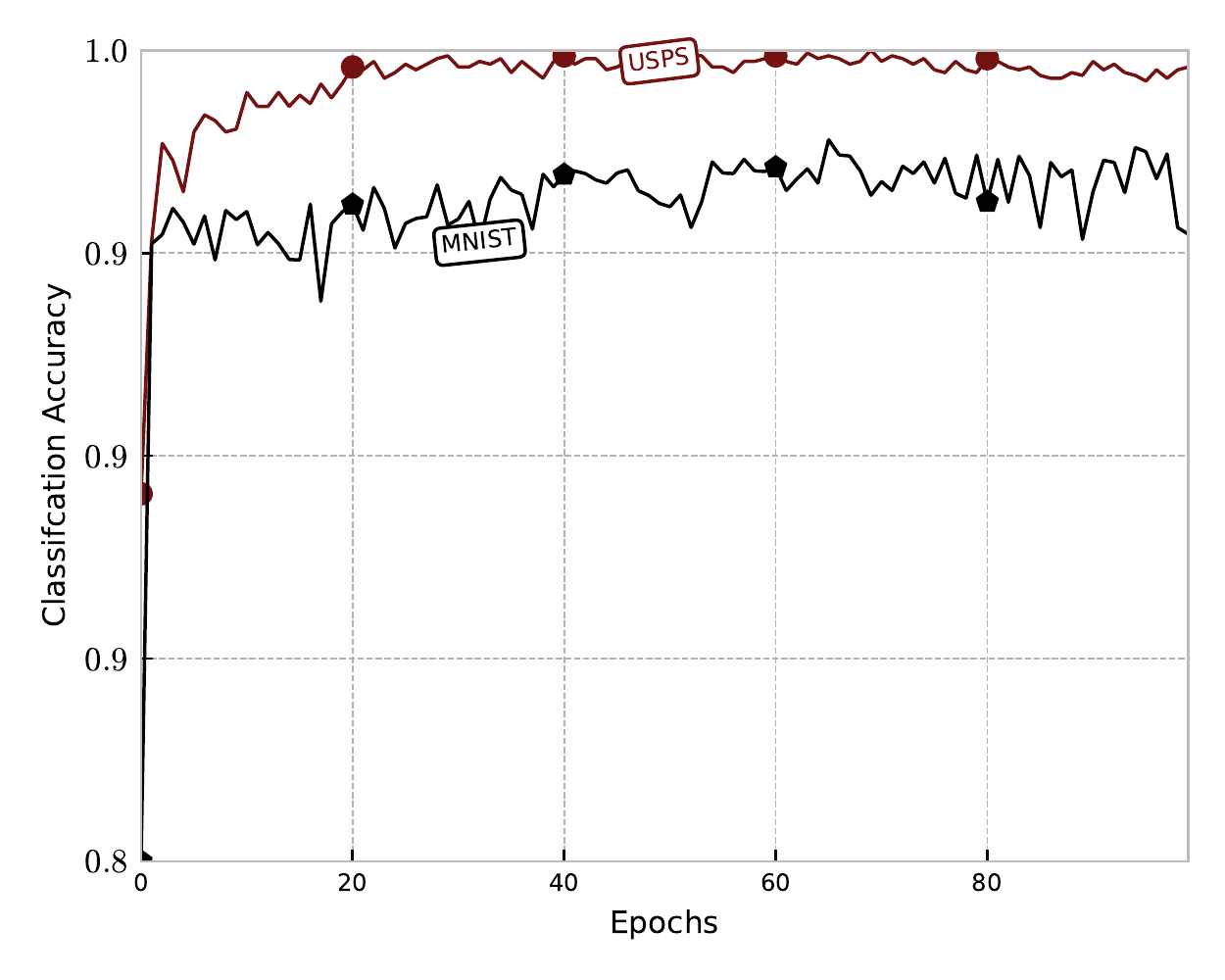}
    \caption{{\bf Optimizing the triplet loss.} (left) the Triplet loss value during the training of the source model on the USPS and MNIST datasets; (right) The classification accuracy obtained on the target datasets.}\label{fig:loss}
\end{figure}

\begin{figure}[!t]
    \centering
       \includegraphics[width=0.5\textwidth]{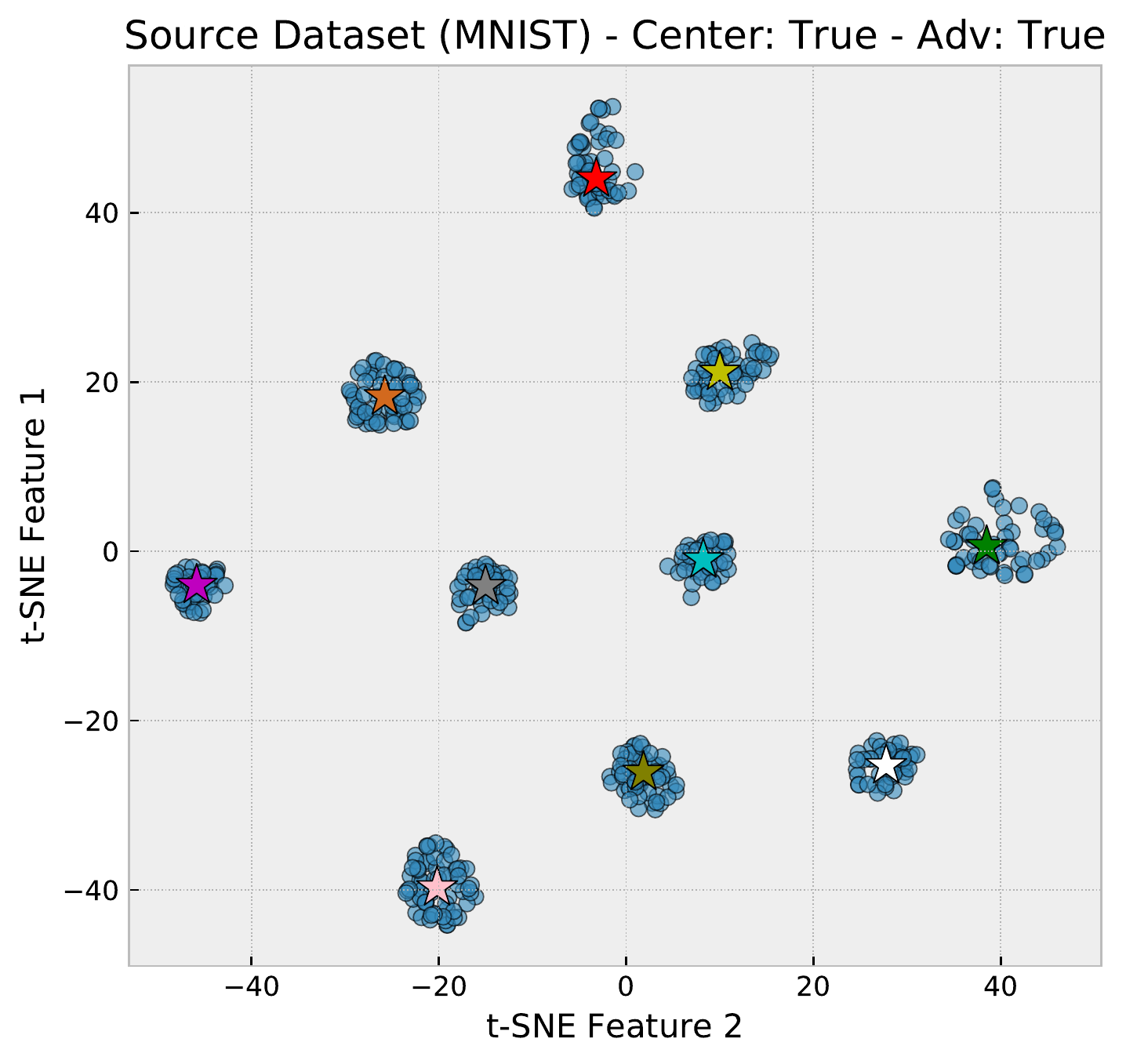}~
        \includegraphics[width=0.5\textwidth]{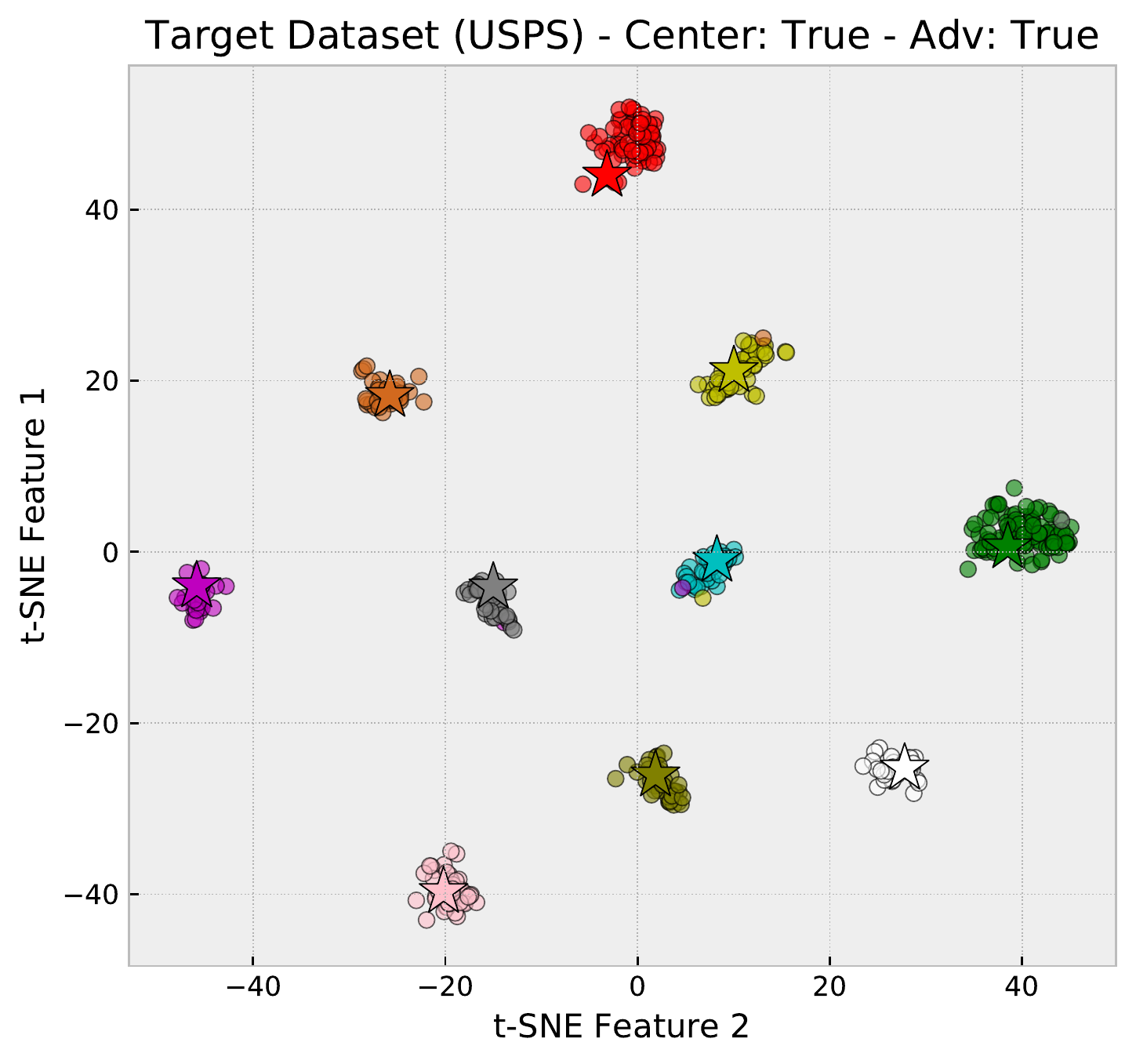}
    \caption{{\bf M-ADDA results.} (left) The t-SNE components of the source embeddings on the MNIST dataset after training the source model. (right) The t-SNE components of the target embeddings of the USPS dataset after training the target model. The stars represent the cluster centers of the source embeddings. The colors represent different labels.}\label{fig:both}
\end{figure}

\begin{table}[t]
\centering
\caption{{\bf Ablation studies}. Impact of the loss terms on the classification accuracy of the target model.}
\begin{tabular}{ |l|c|c| } 
   Method & MNIST $\rightarrow$ USPS & USPS $\rightarrow$ MNIST \\\hline\hline
Center Magnet Only& 0.77 & 0.85 \\\hline
Adversarial Adaptation Only & 0.93 &  0.92\\\hline
M-ADDA & {\bf \;0.98 } & {\bf \;0.97 } \\\hline
\end{tabular}
\label{table:ablation}
\end{table}

To illustrate the performance of our method for the unsupervised domain adaptation task, we apply it on the standard digits dataset benchmark using accuracy as the evaluation metric. We consider 2 domains: MNIST, and USPS. They consist of 10 classes representing the digits between 0 and 9 (we show some digit examples in Figure \ref{fig:digits}). We follow the experimental setup in \cite{2017arXiv170205464T} where 2000 images are sampled from MNIST and 1800 from USPS for training. Since our task is unsupervised domain adaptation, all the images in the target domain are unlabeled. In each experiment, we ran Algorithm \ref{alg:source} for 200 epochs to train our source model. Then, we report the accuracy on the target test set after running Algorithm \ref{alg:da} for 200 epochs.

We also use similar architectures for our models as those in \cite{2017arXiv170205464T}. The encoder module is the modified LeNet architecture provided in the Caffe source code \cite{lecun1998gradient}.  The decoder is a simple linear model that transforms the encoded features into 256-unit embedding vectors. The discriminator consists of 3 fully connected layers: two layers with 500 hidden units followed by the final discriminator output. Each of the 500-unit layers uses a ReLU activation function.

Table \ref{table:results} shows the results of our experiments  on the digits datasets. We see that our method achieves competitive results compared to previous state-of-the-art methods, ADDA \cite{2017arXiv170205464T}. This suggests that metric learning allows us to achieve good results for domain adaptation. Further, Table \ref{table:resultsBig} shows the results of our experiments  using the setup in \cite{bousmalis2016domain,bousmalis2017unsupervised} where the full training set was used for both MNIST and USPS. We see that our method beats recent state-of-the-art methods in the USPS, MNIST domain adaptation challenge. However, it would be interesting to see the efficacy of M-ADDA in more complicated tasks such as the VisDA dataset challenge \cite{peng2017visda}. We show in Fig. \ref{fig:loss} (left) the Triplet loss value during the training of the source model on the USPS and MNIST datasets. Further, Fig. \ref{fig:loss} (right) shows the classification accuracy obtained on the target datasets with respect to the number of epochs. Higher accuracy was obtained for USPS when the model was trained on MNIST, which is expected since MNIST consists of more training examples.

\begin{figure}[t]
    \centering
        \includegraphics[width=0.5\textwidth]{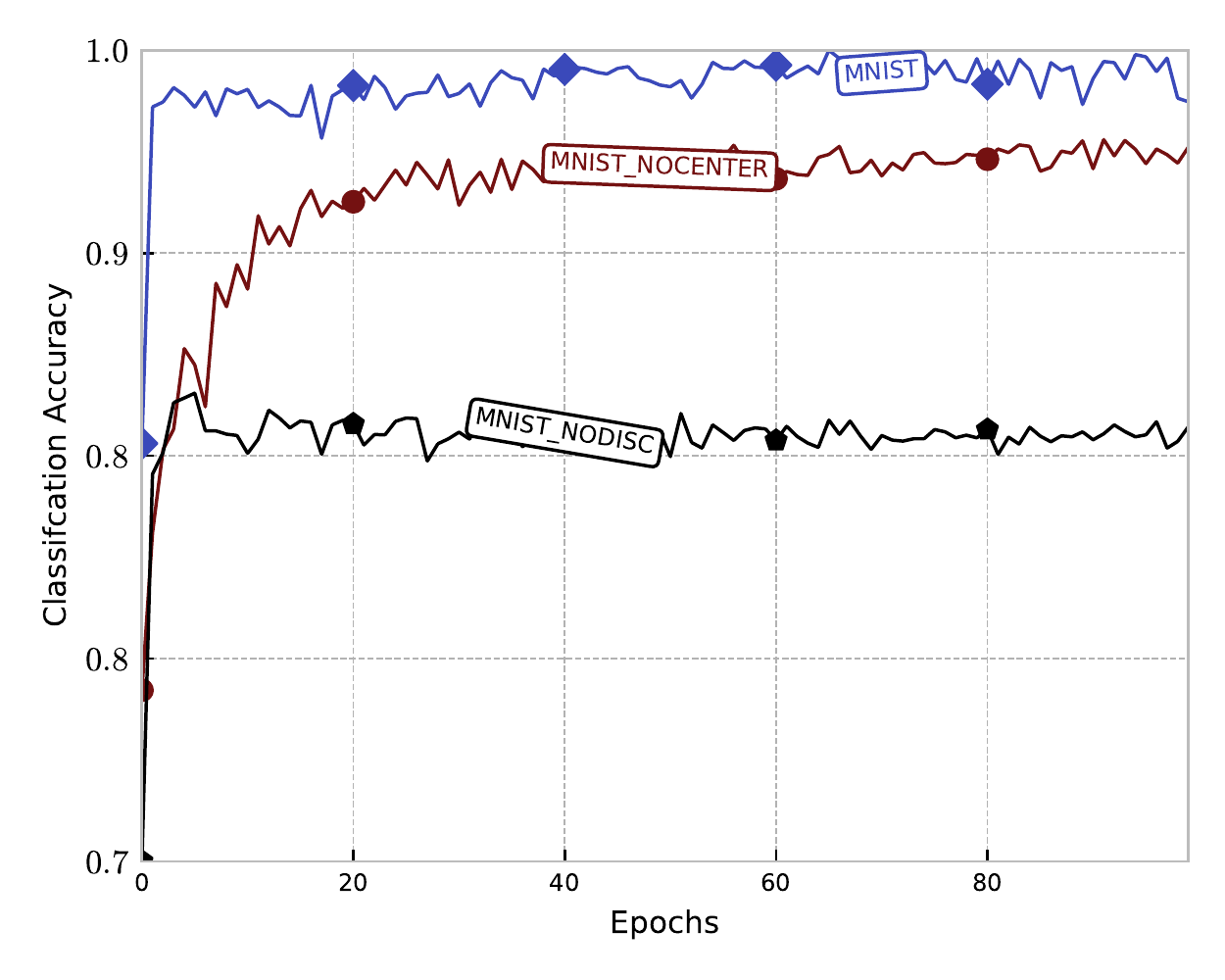}~
        \includegraphics[width=0.5\textwidth]{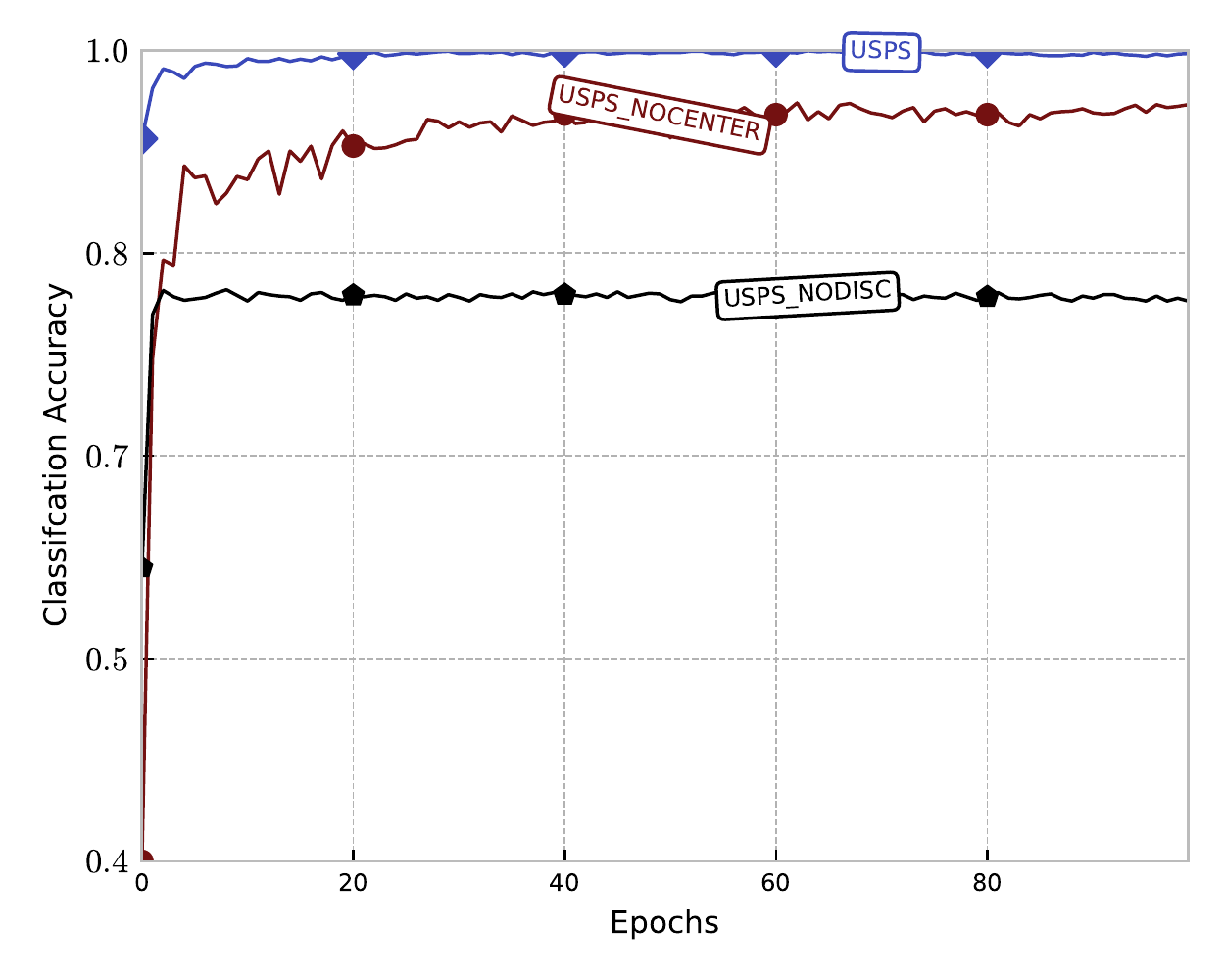}
    \caption{{\bf Ablation studies.} (left) The classification accuracy on MNIST using variations of the loss function (\ref{eq:dacluster}); (right) The classification accuracy on USPS using variations of the loss function (\ref{eq:dacluster}). NOCENTER refers to optimizing Eq. (\ref{eq:da}) only, and NODISC refers to optimizing Eq. (\ref{eq:cluster}) only. The blue lines refer to the result of optimizing Eq. (\ref{eq:dacluster}).}\label{fig:acc}
\end{figure}

\begin{figure}[!t]
    \centering
        \includegraphics[width=0.5\textwidth]{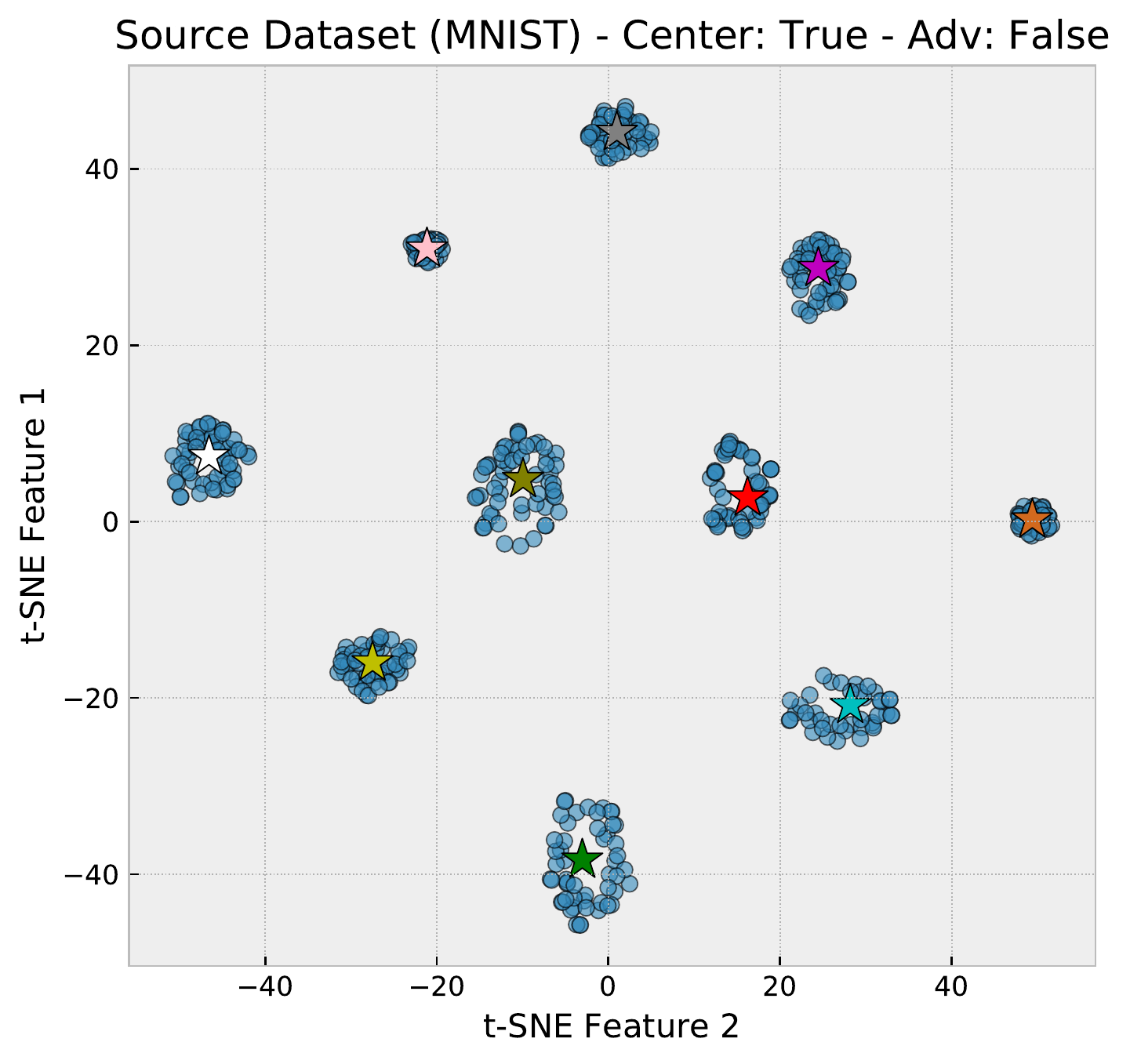}~
        \includegraphics[width=0.5\textwidth]{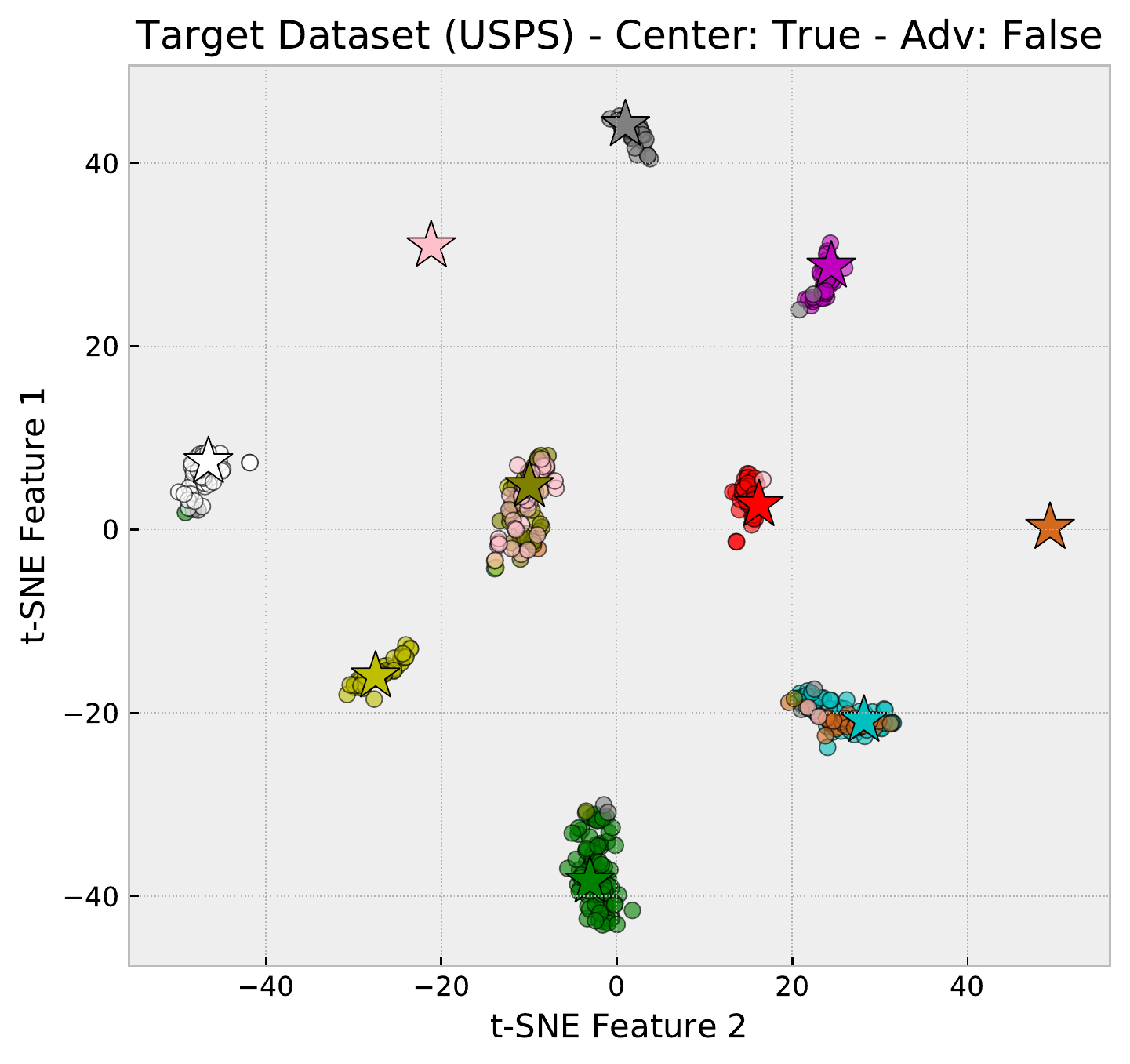}
    \caption{{\bf Center magnet optimization only.} The stars represent the cluster centers of the source embeddings.}\label{fig:centeronly}
\end{figure}
\begin{figure}[!t]
    \centering
        \includegraphics[width=0.5\textwidth]{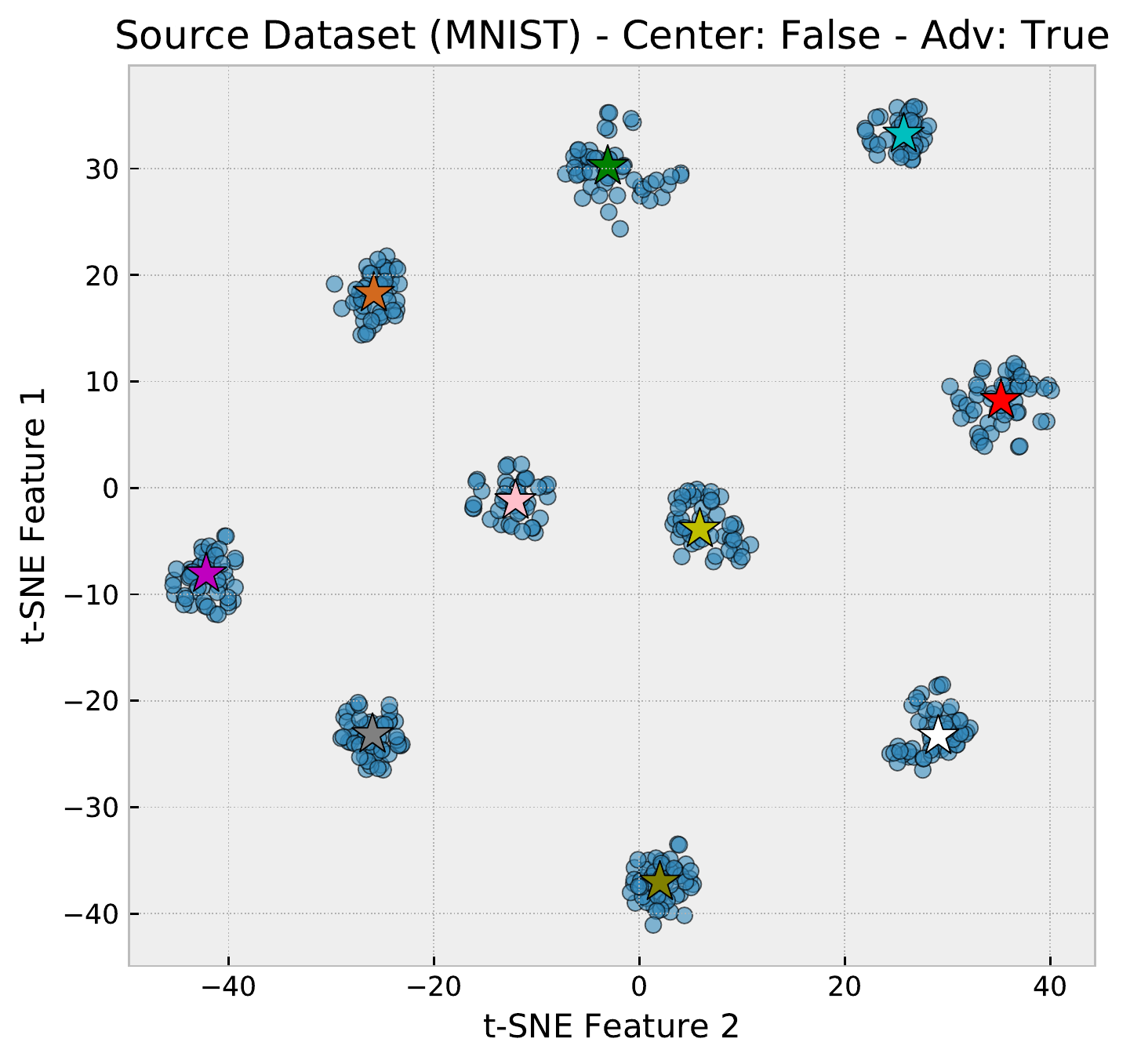}~
        \includegraphics[width=0.5\textwidth]{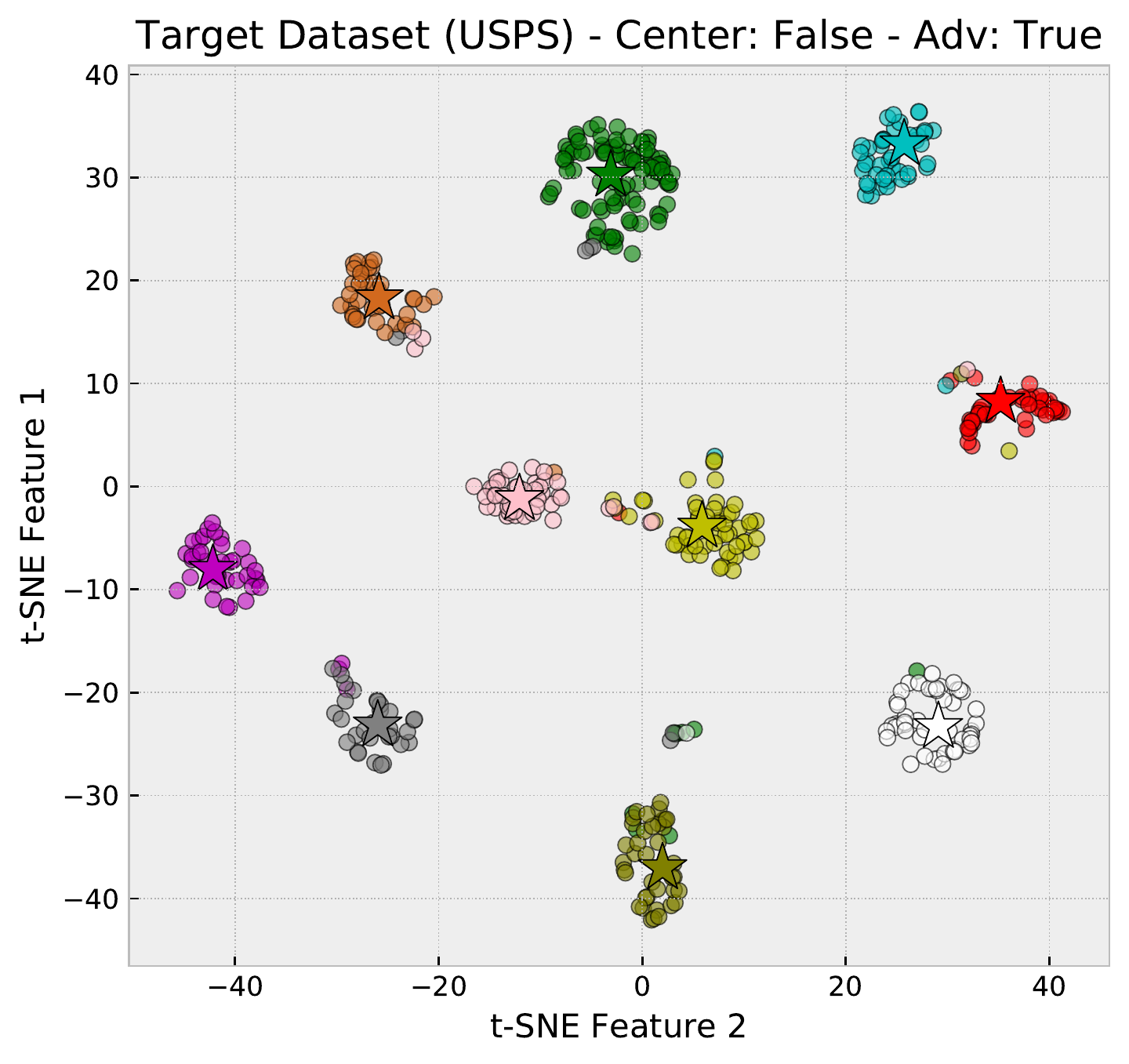}
    \caption{{\bf Adversarial optimization only.} The stars represent the cluster centers of the source embeddings.}\label{fig:daonly}
\end{figure}

In Table \ref{table:ablation}, we compare between two main variations for training the target model. Center Magnet only updates the target model using only Eq. (\ref{eq:cluster}); therefore, it ignores the adversarial training part of Eq. (\ref{eq:da}). Using Center Magnet only to train the target model results in poor performance. This is expected since the performance highly depends on the initial clustering. We see in Fig. \ref{fig:centeronly} (right) that several source cluster centers (represented as stars) contain samples corresponding to different labels. For example, the samples with the pink label are clustered with those of the green label. Similarly, those with the orange label are clustered with those of the teal label. This is expected since the target model is encouraged to move the embeddings to the nearest cluster centers without having to match the extracted feature distributions between the source and target datasets.

Using only the adversarial adaptation loss improves the results significantly, since having the extracted features distribution between the source and target similar is crucial. However, we see in Fig. \ref{fig:daonly} (right) that some samples are far from any cluster center which makes their class labels ambiguous. Namely, the pink and yellow samples that are in the center between the yellow and pink cluster centers. To address these ambiguities, the center magnet loss helps the model to regularize against them. As a result, we see in Fig. \ref{fig:both} (right) that better clusters are formed when we optimize the whole loss function defined in Eq. \ref{eq:dacluster}. This suggests that M-ADDA has strong potential in addressing the task of unsupervised domain adaptation.

\section{Conclusion}\label{sec:conclusion}
We propose M-ADDA, which is a metric-learning based method, to address the task of unsupervised domain adaptation. The framework consists of two main steps. First, a triplet loss is used to pre-train the source model on the source dataset. Then, we adversarialy train a target model to adapt the distributions of its extracted features to match those of the source model. In parallel, we optimize a center magnet loss to regularize the output embeddings of the target model so that they form clusters that have similar structure as that of the source model's output embeddings. We showed that this approach can perform significantly better than ADDA \cite{2017arXiv170205464T} on the digits adaptation dataset of MNIST and USPS. For future work, it would be interesting to apply these methods on more complicated datasets such as those in the VisDA challenge.

\bibliographystyle{splncs04}
\bibliography{refs}

\end{document}